\documentclass[11pt,a4paper]{article}
\usepackage[hyperref]{eacl2021}
\usepackage{times}
\usepackage{latexsym}

\usepackage{microtype}

\aclfinalcopy 

\usepackage{graphicx}
\usepackage{amsmath}
\usepackage{booktabs}
\usepackage{multirow}
\usepackage{mathtools}

\title{Hierarchical Transformer for Multilingual Machine Translation}

\author{
  Albina Khusainova \\
  Innopolis University, \\
  Innopolis, Russia \\
  \texttt{a.khusainova@innopolis.ru} \\\And
  
  Adil Khan \\
  Innopolis University, \\
  Innopolis, Russia \\
  \texttt{a.khan@innopolis.ru} \\\AND
  
  Adín Ramírez Rivera \\
  University of Campinas \\
  Campinas, Brazil \\
  \texttt{adin@ic.unicamp.br} \\\And
  
  Vitaly Romanov \\
  Innopolis University, \\
  Innopolis, Russia \\
  \texttt{v.romanov@innopolis.ru}
  }

\date{}

\begin{document}
\maketitle
\begin{abstract}

The choice of parameter sharing strategy in multilingual machine translation models determines how optimally parameter space is used and hence, directly influences ultimate translation quality. Inspired by linguistic trees that show the degree of relatedness between different languages, the new general approach to parameter sharing in multilingual machine translation was suggested recently. The main idea is to use these expert language hierarchies as a basis for multilingual architecture: the closer two languages are, the more parameters they share.


In this work, we test this idea using the Transformer architecture and show that despite the success in previous work there are problems inherent to training such hierarchical models. We demonstrate that in case of carefully chosen training strategy the hierarchical architecture can outperform bilingual models and multilingual models with full parameter sharing.

\end{abstract}

\section{Introduction}

Machine translation (MT) today is gradually approaching near-human quality; however, this holds true only when massive parallel corpus is available. As for the low-resource machine translation, the main way to improve it is to use additional data such as monolingual texts or parallel data in other languages. One of the ways to use the latter is to build a multilingual model: instead of training separate models for each translation direction, multiple parallel corpora can be combined to train a single model where languages can share some parameters to help each other learn. Given that languages have a lot in common, a properly organized parameter sharing strategy could compensate for the lack of training examples in low-resource pairs.

Exploiting language relatedness can substantially improve translation quality \cite{Tan2019MultilingualNM}.
The question is which architectural setup allows getting the most benefit from between-language commonalities. While there are different approaches to this problem, such as full parameter sharing \citep{johnson-etal-2017-googles} or shared encoder side with language-specific decoders \citep{dong-etal-2015-multi}, we find the recent approach by \citet{azpiazu2020framework} promising because it accounts for the degree of relatedness between languages in a multilingual model in a systematic way. 

\begin{figure*}[tb]
 \centering
  \includegraphics[width=\textwidth]{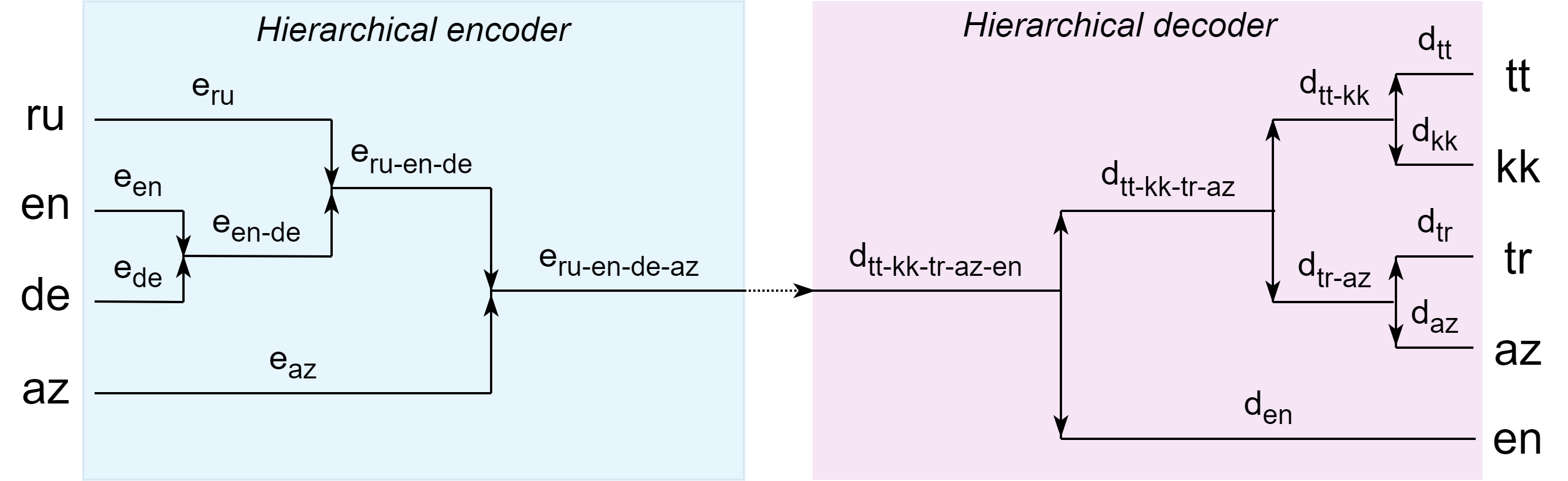}
 \caption{\label{multi} A high-level view of a sample multilingual hierarchical model. Horizontal lines denote encoder/decoder blocks, vertical lines and arrows show points of merging/splitting, and subscripts clarify which languages share a particular encoder/decoder. BCP-47 language codes are given here and below \cite{rfc5646}.}
\end{figure*}

Linguistic trees organize languages in hierarchies by the degree of kinship, and the same approach can be applied to multilingual machine translation models. The idea is to organize both encoder and decoder in a hierarchical fashion, reflecting the degree of relatedness between languages, such that the most related languages share the largest number of parameters. Figure~\ref{multi} shows the outline of such model (will be explained in more detail in section~\ref{sec:model_descr}).

Unlike \citet{azpiazu2020framework}, who used Long Short Term Memory (LSTM) recurrent neural network as a basic architecture, we implemented this idea using the state-of-the-art Transformer~\cite{transformer} architecture. Our experiments demonstrated some persistent training problems associated with the hierarchical architecture---namely, the model is prone to early overfitting in low-resource directions leading to low translation scores. Given that one of the main goals of introducing multilingual models in general and the hierarchical model in particular is improving translation accuracy for low-resource directions, this problem is critical and needs a solution.

These are the main contributions of this work:

1. Testing the hierarchical multilingual translation model using the Transformer architecture;

2. Identifying and analyzing problems related to the hierarchical nature of the model---namely, overfitting in low-resource directions;

3. Suggesting and testing a number of solutions, which could be summarized as various forms of regularization.


\section{Related Work}
\label{sec:related}
In this section, we review the existing approaches to parameter sharing in multilingual MT models to contrast them with the hierarchical approach.  

Probably, the simplest form of parameter sharing was introduced by \citet{johnson-etal-2017-googles}, where all parameters are shared and identifier tokens are used to distinguish between languages. While being easy to implement, such model may require more resources to capture relationships between languages, as no architectural hints are given.

\citet{dong-etal-2015-multi} views the translation problem as a multi-task one, with a shared encoder and separate decoders for each target language. Here the potential knowledge sharing between target languages is not considered.

\citet{firat-etal-2016-multi} and \citet{wang-etal-2019-compact} introduce shared parameters between the encoder and decoder. This is an interesting idea that can be considered in the future.



\citet{sachan-neubig-2018-parameter} propose a one-to-many model, where parameters are partially shared between the multiple decoders. This is similar in spirit to the hierarchical approach; however, instead of having shared parts between individual decoders, the idea is to build a hierarchy of decoders.


Many other recent works, such as \citet{Bapna2019SimpleSA} and \citet{Fan2020BeyondEM} add language-specific components/layers to the decoder side of a model.

We observe that in these works languages on the encoder/decoder side of multilingual models are treated the same, regardless of the degree of their relatedness. In fact, \citet{Fan2020BeyondEM} do group languages by vocabulary overlap, nevertheless, in their model close languages may end up being in different groups. In the hierarchical model, however, the number of shared parameters directly depends on the degree of kinship between languages. 

\section{Approach}
\label{sec:approach}

Using trees to model genealogical relationships between different languages has historically been a common approach in linguistics \cite{schleicher1853ersten}. Although this is not the only model, and there exist alternatives such as wave model \cite{schmidt1872verwantschaftsverhaltnisse}, tree structure naturally suits for depicting origination of languages one from another over time.


This expert knowledge of relationships between languages can be utilized when building a multilingual MT model. The current trend in natural language processing is to make models learn linguistic rules and patterns on their own, without explicitly guiding them. However, in this case, when supervision does not really cost anything and language relatedness information is readily available, this expert knowledge can let models train faster and manage parameter space better.


\citet{belinkov-etal-2017-evaluating} found out that different encoder levels specialize in different language aspects: lower-level representations better capture part-of-speech tags, and higher-level ones are better at grasping semantics.

Another recent observation by \citet{kudugunta-etal-2019-investigating} showed that in multilingual MT models with a single shared encoder, representations of different languages cluster by language families, and as they move up the encoder, similarity between source languages increases.

These facts taken together suggest that the encoder tries to find a common representation of different languages that initially cluster together by language families. As they move up the encoder, a model finds intermediate representations that smooth out the dissimilarities on different levels - morphological, syntactic, semantic. 

\begin{figure}[tb]
 \centering
  \includegraphics[width=0.49\textwidth]{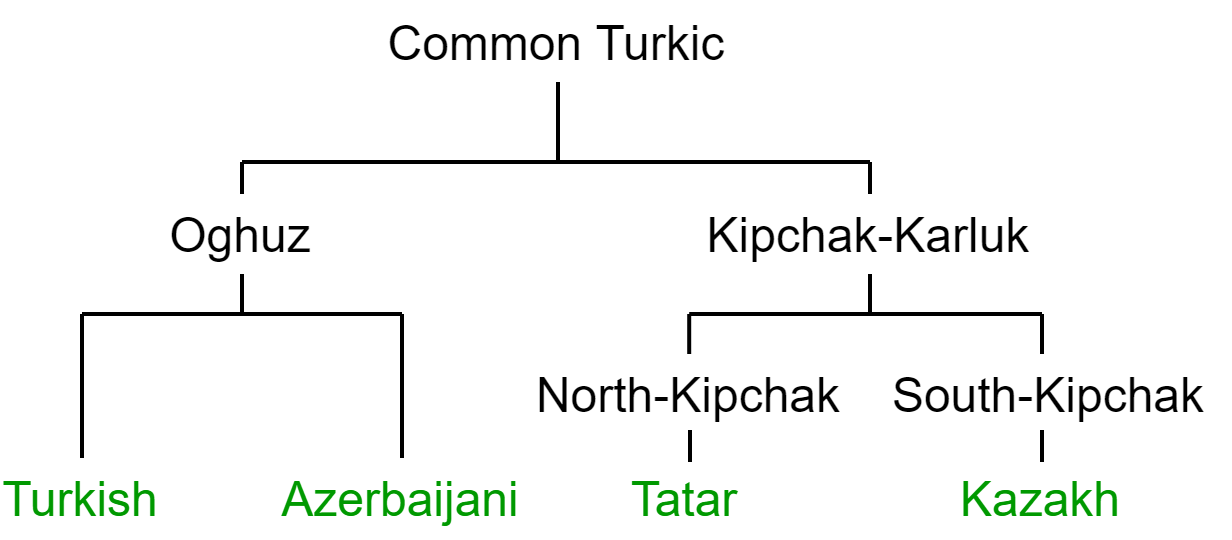}
 \caption{\label{turk} A fragment of the phylogenetic tree for Turkic languages \cite{turkic_tree}.}
\end{figure}

This closely resembles the structure of phylogenetic trees, where connections on the bottom level mean the greatest linguistic similarity between languages, and connections on the highest level of a hierarchy indicate that languages are far from each other. See the example of such tree in Figure~\ref{turk}. If we organize the architecture of a multilingual MT model according to language relationships in a phylogenetic tree, this will allow sharing parameters between languages on appropriate levels. 

For example, we can take two very close languages, whose vocabularies overlap significantly. Since these languages are highly related, we assume that it will not take long to reduce them to the same representation. So, we ``combine" them early on, i.e., introduce shared parameters for these languages in the first layers of the encoder. The third language, let us suppose, comes from the same family, but many base words are different. Still, sentence structure remains the same as in the first two languages. Therefore, we combine this language with the first two in the later stages (parameter sharing starts deeper in encoder layers), and so on. This parameter sharing strategy allows the economic utilization of parameter space and has the potential to lead to better, i.e., source-language-independent representations of different languages on the output level of the encoder.

\subsection{Model Description}
\label{sec:model_descr}

The principal idea is to organize languages on the encoder and decoder sides according to their linguistic similarity. Namely, each side of the model is built as a hierarchy that corresponds to how languages connect in phylogenetic trees: the closer two languages are, the more parameters they share. 

The example of the hierarchical model is given in  Figure~\ref{multi}. There are four source and five target languages connected through the chain of hierarchically organized encoders and decoders. Let us consider the encoder side. All source languages have their own respective encoders $e_{ru}$, $e_{en}$, $e_{de}$, $e_{az}$, and parameters of these encoders are not shared with any other language since they are intended to learn language-specific features. Deeper in the model, the outputs of some encoders are stacked together and passed to shared encoders (connections are denoted with arrows). These encoders ($e_{en-de}$, $e_{ru-en-de}$) are shared among two or more languages so that they can capture knowledge common to these languages. Finally, there is the last encoder $e_{ru-en-de-az}$ that is shared by all source languages and that combines the outputs of all remaining encoders together.

This architecture is designed to enable knowledge sharing between different languages on different levels. We see that English and German are connected first, as they both come from the Germanic branch of the Indo-European language family. Further in the model, the Russian language mixes in, since it belongs to a different branch of the same language family. Later on, we join these Indo-European languages with Azerbaijani, which comes from an entirely different Turkic language family. Logic on the decoder side is analogous---most dissimilar languages split first.

We suggest having the same number of parameters along any path from source to target. For example, if we compare ru-tt and az-en translation paths in Figure~\ref{multi}, although the latter has considerably less shared parameters, their total number should be the same for both paths. That is why some encoder blocks are longer than others, pointing that, for instance, the number of parameters in $e_{az}$ should be the same as there are cumulatively in $e_{ru}$ and $e_{ru-en-de}$. 

Overall, we can say that the model adheres to the framework by \citet{azpiazu2020framework}, but there are important differences. We use a different underlying model---Transformer instead of LSTM, we do not limit the number of layers and do not prune language families. There are also differences in training procedure which will be explained in section~\ref{sec:model_training}. And, most importantly, we identify the problem specific to training hierarchical models and suggest the improved training techniques for such models, as will be described in section~\ref{sec:results}.

\section{Experiments}
\label{sec:experiments}

To test the hierarchical approach to parameter sharing we experimented with several hierarchical setups in multilingual models. Unlike \citet{azpiazu2020framework} who only test complex multi-source multi-target models, we decided to start with simple setups when the hierarchy is just on the one side of the model (encoder or decoder) and then increase the complexity to the general case.

\begin{figure}[tb]  
 \centering
  \includegraphics[width=0.5\textwidth]{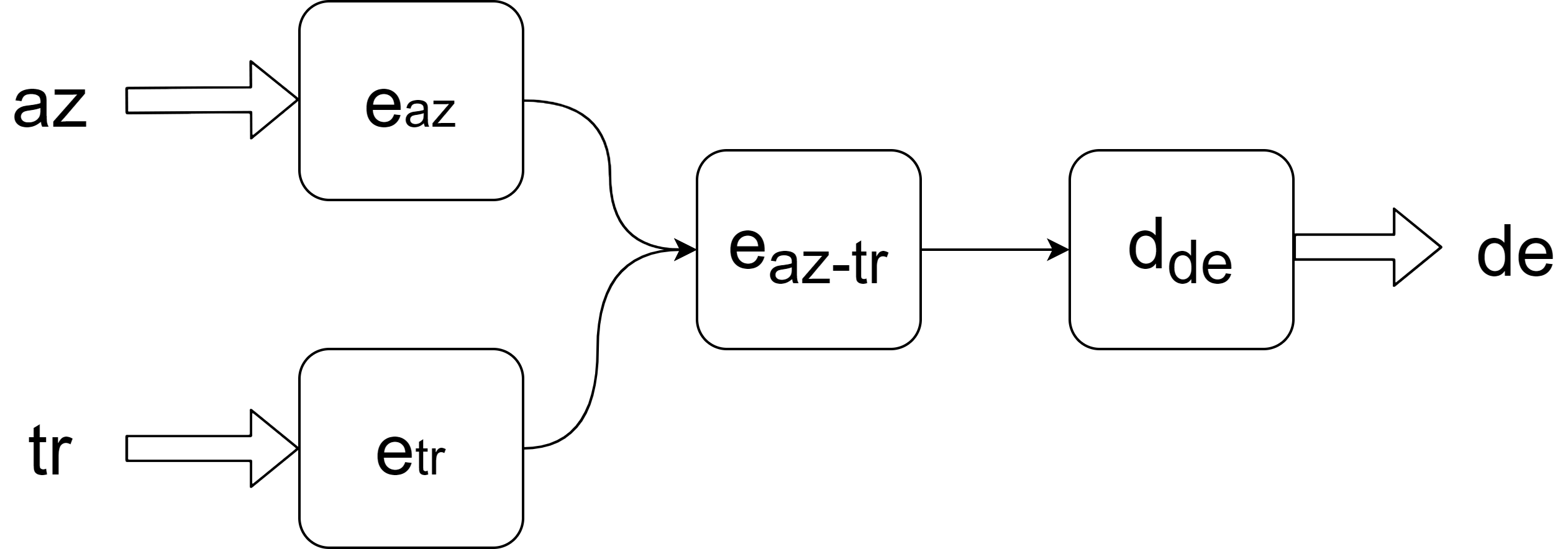}
 \caption{\label{hie_aztr} The example of a simple hierarchical model with two source and one target language. Rounded rectangles denote encoder/decoder blocks.}
 \end{figure}

\begin{figure*}[tb] 
 \centering
  \includegraphics[width=0.8\textwidth]{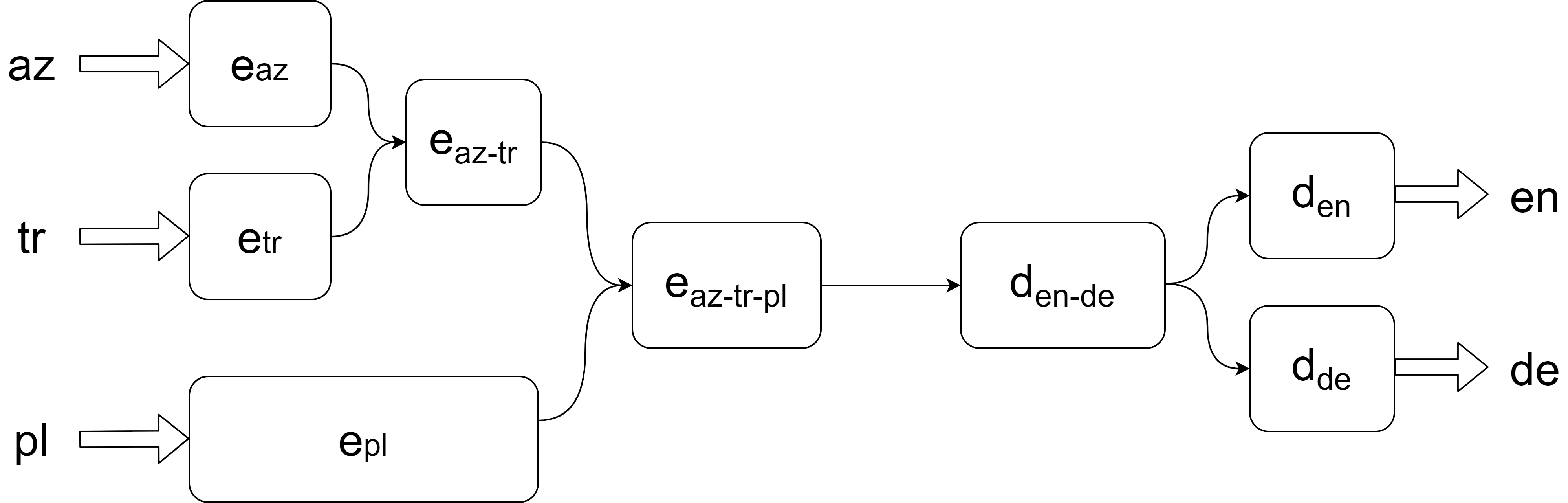}
 \caption{\label{hie_aztrpl-ende} The example of a general multi-source multi-target hierarchical model.}
\end{figure*}

We tested the hierarchical model's performance in three different cases: 

\begin{enumerate}
    \item  Simple case with two related source languages and one target language, Figure~\ref{hie_aztr};
    \item Simple case with one source language and two related target languages;
    \item General case with several languages of different degree of relatedness both on the source and target side, Figure~\ref{hie_aztrpl-ende}.
\end{enumerate}

For each of these cases we trained two hierarchical models using different language sets. Further, when we identified specific training problems of hierarchical models (described in section~\ref{sec:problem}), we suggested and applied two variations to each hierarchical model.


Each hierarchical model is compared to two baselines: bilingual models and multilingual model with full parameter sharing (will be also called \textit{full sharing model} later). Comparing to bilingual models trained on the same parallel corpora helps to understand whether languages can learn from each other in the hierarchical model. Whereas contrasting hierarchical models to full sharing models, with single shared encoder and decoder for all languages, as in \citet{johnson-etal-2017-googles}, answers the question about the use of explicitly introducing the hierarchy into the model.

In every multilingual model we trained there are low-resource and high-resource directions, and we explored whether low-resource ones can learn from high-resource ones, and which model structure better fits for this purpose.

\subsection{Data}
We used four parallel corpora, all coming from the same JW300 dataset \cite{agic-vulic-2019-jw300} to enable fair comparison within the same domain: 

\begin{itemize}
\itemsep0em
    \item Turkish-German (tr-de), 500k sentences;  
    \item Azerbaijani-German (az-de), 110k;
    \item English-Polish (en-pl), 500k;
    \item German-Polish (de-pl), 110k.
\end{itemize}

Turkish and Azerbaijani languages are related and come from the Oghuz sub-branch of the Turkic language family. English, German, and Polish all belong to the Indo-European language family, however, English and German are more related as they both come from the Germanic branch, while Polish is under the Balto-Slavic branch. 

For training simple models with a hierarchy on one side (case 1 and case 2) we used tr-de and az-de corpora together (resulting models can be denoted as az-tr$\rightarrow$de (Figure~\ref{hie_aztr}) and de$\rightarrow$az-tr). Similarly, en-pl and de-pl corpora  resulted in en-de$\rightarrow$pl and pl$\rightarrow$en-de models.

For general hierarchical models (case 3) all four corpora were used together and resulted in 2 models: az-tr-pl$\rightarrow$en-de (Figure~\ref{hie_aztrpl-ende}) and en-de$\rightarrow$az-tr-pl. 

The sizes of corpora are different: tr-de and en-pl are considered as high-resource language pairs, and az-de and de-pl are low-resource pairs. 

All corpora were filtered by maximum sentence length of $40$ BPE tokens \cite{sennrich2016bpe}.

\subsection{Model and Training}
\label{sec:model_training}

Our multilingual machine translation system implementation is based on the Transformer architecture \cite{transformer} with the reduced number of parameters:

\begin{itemize}
\itemsep0em
    \item d\_model $= 128$
    \item dff $= 512$
    \item num\_heads $= 8$
    \item dropout\_rate $= 0.1$
\end{itemize}

For all hierarchical and bilingual models there are 6 layers on the path from any source to any target language. Thus, any bilingual model consists of 3 encoder and 3 decoder layers. Case 1 model (with hierarchical encoder) consists of 1 layer in each of the individual encoders, 2 layers in the shared encoder, and 3 layers in the decoder. Adding more languages in the general hierarchical model results in more overall layers, but there are still 6 layers between any pair of source and target languages. 

To make the models with full parameter sharing comparable to hierarchical ones, we keep the overall number of layers in the encoder and decoder the same between the models. For example, the full parameter sharing model trained for comparison with case 1 hierarchical model has 4 encoder and 3 decoder layers (corresponds to 1+1+2 and 3). However, when we came to the general case with more layers, the full sharing model could not learn anything. So, for en-de$\rightarrow$az-tr-pl doubling batch size solved the problem, and for az-tr-pl$\rightarrow$en-de we had to decrease the number of layers in the encoder from 6 to 4.  

\citet{azpiazu2020framework} did not mention whether they kept the sizes of the hierarchical model and their baselines comparable, which makes it difficult to analyze the results.

Since training corpora are of different sizes, we oversampled low-resource corpora when training hierarchical models. For full parameter sharing models, however, it turned out that oversampling hurts their performance, so we did not apply it.

As for the training procedure, one way to train the hierarchical model is to alternate between all translation directions in a system (as \citet{azpiazu2020framework}). In this case, the possible concern is that the model parameters may start to oscillate between these directions. Therefore, we decided to simultaneously feed data from all source languages to the respective encoders, stack representations at points where specific encoders merge into shared ones, and pass them through the chain of decoders down to individual decoders. We trained all models with batch size 128 until convergence (for 50 epochs maximum) and report best BLEU scores reached over epochs. When training hierarchical models, we stack full-size ($128$) batches of different language pairs in the encoder and split them in the decoder. 

To facilitate knowledge sharing in multilingual models we use shared vocabularies. 

\subsection{Results and Analysis}
\label{sec:results}

In this section, we present and discuss the results of the experiments.

\subsubsection{Evaluating Hierarchical Model}

To evaluate the hierarchical approach, we used two baselines: bilingual and full sharing models. We looked at several metrics. First, we calculated the average difference in BLEU scores between multilingual and bilingual models. To do so, we averaged across total 16 translation paths in all trained full sharing models, and, separately, in all hierarchical models. Second, we divided these translation paths to low-resource (8) and high-resource ones (8), and computed average BLEU scores difference across high-resource directions only and, third, across low-resource directions only.

\begin{figure*}[tb]
 \centering
  \includegraphics[width=0.9\textwidth]{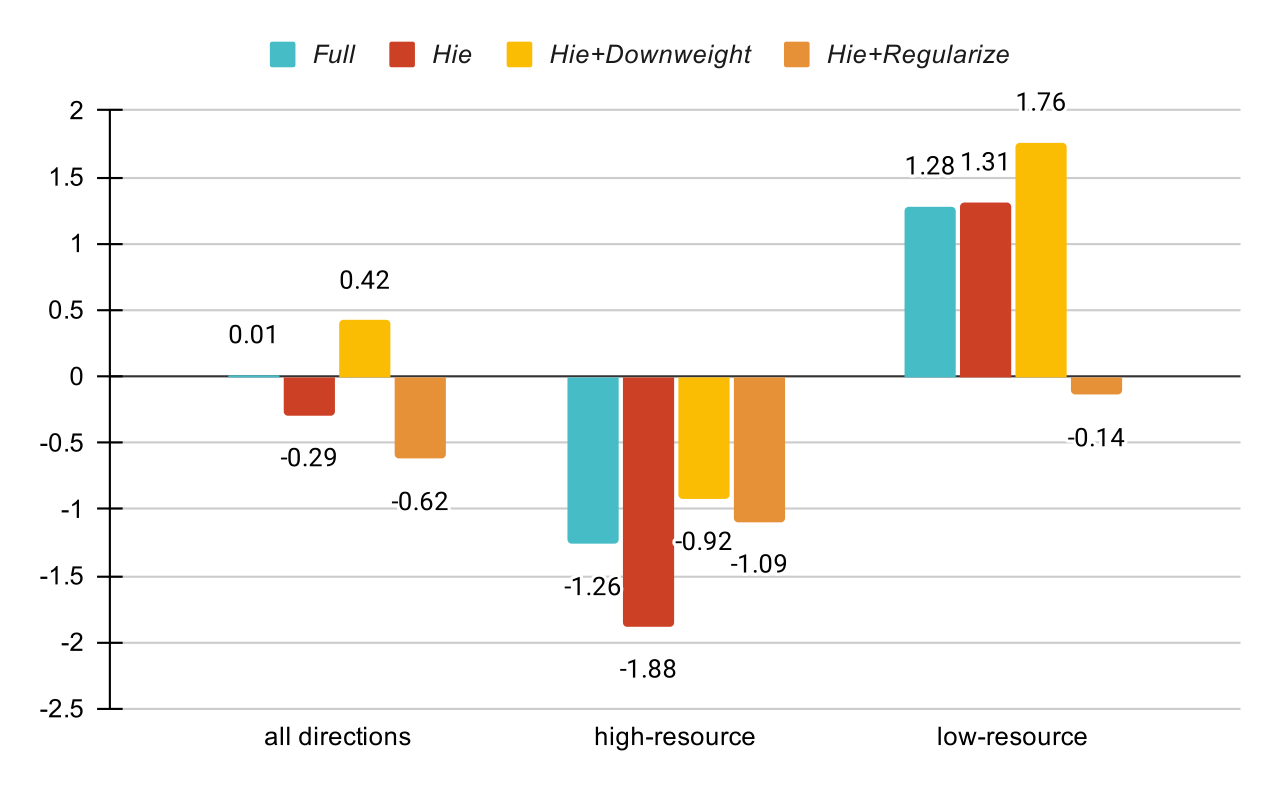}
 \caption{\label{chart}  Difference in BLEU scores between multilingual models and bilingual models. Zero level represents average bilingual models' score. Positive values mean improvement over bilingual baseline, negative values, respectively, signify decrease. \textit{Full} denotes models with full parameter sharing, \textit{Hie} stands for hierarchical models, \textit{Hie+Downweight} is for hierarchical models trained by down-weighting low-resource samples, and \textit{Hie+Regularize} is for hierarchical models regularized using high-resource samples.}
\end{figure*}

This information is visualized in Figure~\ref{chart}. The height of the bars reflects the magnitude of the difference, and their direction indicates whether scores improve or degrade, compared to a bilingual baseline. For now, we are only interested in full sharing (Full) and basic hierarchical (Hie) models. 

On the left of the Figure~\ref{chart} there are average differences in scores, and we see that full sharing models perform on average almost the same as bilingual ones, and hierarchical are even slightly worse (-0.29). However, the comparison becomes more insightful if we look at high-resource and low-resource directions separately. Although on average full sharing models may seem to behave the same as bilingual, now it is clear that this actually happens because low-resource directions improve, learning from related parallel data (+1.28), and high-resource directions degrade (-1.26).

Exactly the same pattern is observed for hierarchical models: low-resource directions do benefit from multilinguality (+1.31), and high-resource directions are hurt by it (-1.88), even more than in full sharing models.

So, we see two problems here: degradation of high-resource directions in multilingual models in general; and overall low performance of hierarchical models, compared to full sharing models. The first problem is a broad one, it was observed in earlier works on multilingual MT models \cite{firat-etal-2016-multi} and thus is outside of the scope of this paper. However, the second problem is surprising, since this violates our assumption about the usefulness of the hierarchical organization of a multilingual model, and hence, needs an investigation.

It would be interesting to compare these results with \citet{azpiazu2020framework}, however, the way they present their findings does not let us do it. They average BLEU scores across unique source languages, and there is no strict division to high-resource and low-resource pairs. According to their results, the hierarchical model performs better than all other baselines, including bilingual and full sharing ones. However, in the GlobalVoices dataset \cite{TIEDEMANN12.463} they used the absolute majority of pairs are very low-resource. Thus, even if high-resource pairs' scores degrade in multilingual models, it would not be seen if for the majority of low-resource pairs scores do improve. The ``bad" high-resource score will simply be lost among low-resource improvements. We cannot assert this is the case, but it certainly can be the case that high-resource pairs' scores in their hierarchical models also degrade. This question could have been clarified in Figure~6 of their paper, where they group language pairs by corpus size, but, unfortunately, they decided to omit results for language pairs with relatively big corpus size (more than 100k). 

Similarly, it also can be the case that high-resource pairs in their full sharing models (called ``one-to-one" there) have higher scores than in hierarchical models.

Overall, it is hard to compare the results in this high/low-resource aspect because settings are very different. Namely, what is low-resource here (100k) could be considered high-resource there. Also, the distribution of high-resource and low-resource directions is different: in our case, half of the directions in multilingual models are high-resource, in their case, they constitute a minority.

So, to sum it up, the identified problems can be the case in \citet{azpiazu2020framework} too, so, our following findings may benefit the overall hierarchical framework they suggested.

\subsubsection{Improving Hierarchical Model}
\label{sec:problem}

The problem we identified and decided to explore here is that the hierarchical model does not outperform the full sharing one as it should have been according to our assumption. The hierarchical idea does not work as expected---it shows almost the same improvements for low-resource pairs and greater degradation for high-resource ones.

\begin{figure*}[tb]
   \centering
  \begin{minipage}[b]{0.45\textwidth}
    \includegraphics[width=\textwidth]{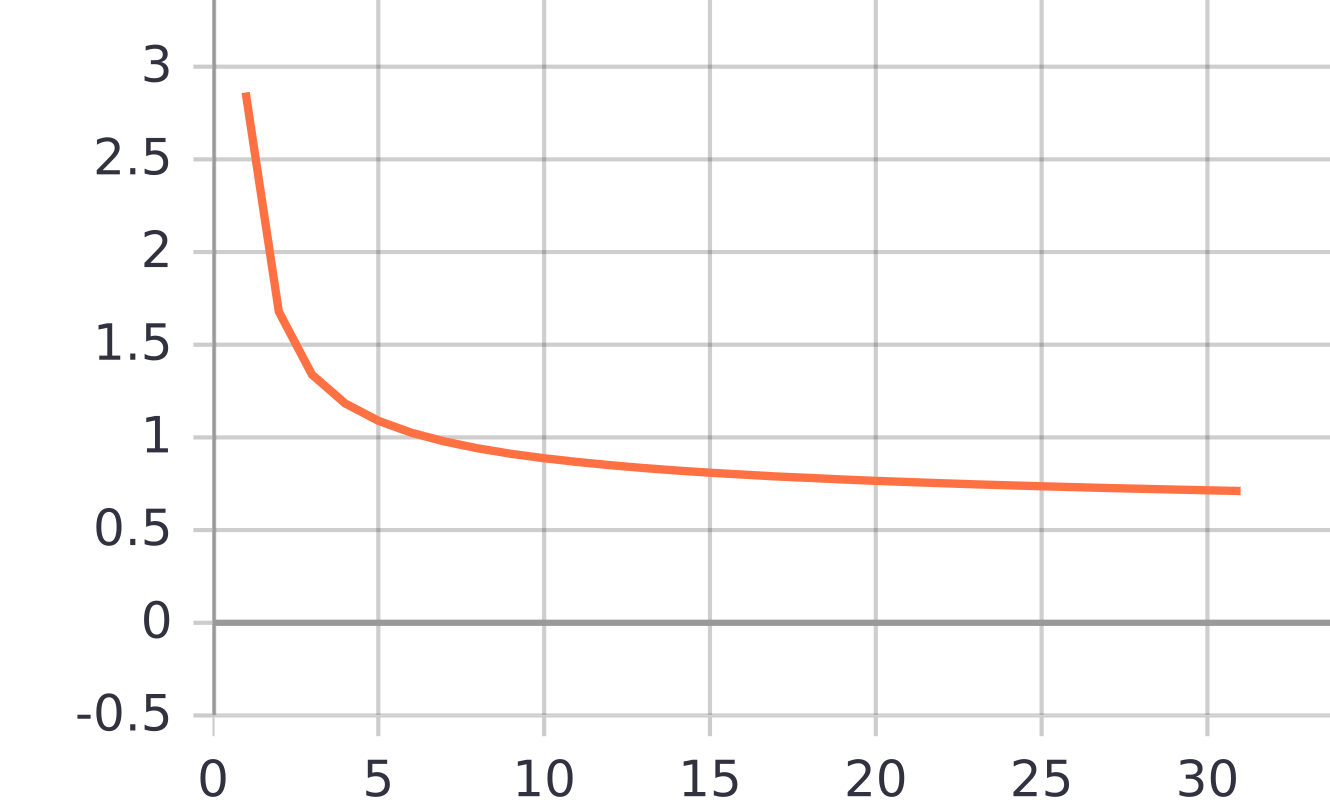}
  \end{minipage}
  \hfill
  \begin{minipage}[b]{0.45\textwidth}
    \includegraphics[width=\textwidth]{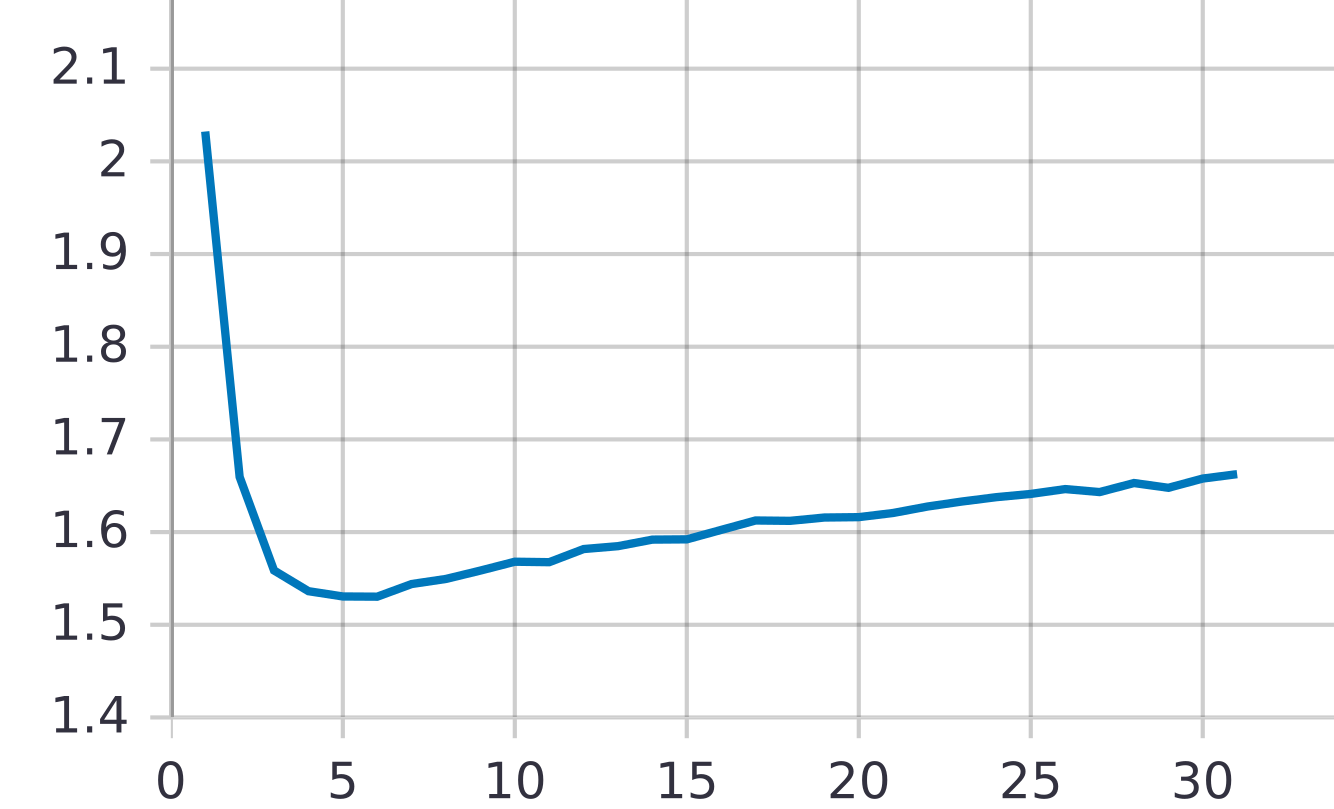}
  \end{minipage}
  \caption{\label{graphs}The example of overfitting in hierarchical models. Train \textit{(left)} and validation \textit{(right)} losses for low-resource pl-de pair when training pl$\rightarrow$en-de hierarchical model.}
\end{figure*}

In an attempt to solve this problem we investigated training dynamics (train and validation losses) for both model types. It turns out there is a persistent problem with training hierarchical models: for all trained models there is a clear overfitting for low-resource pairs, see Figure~\ref{graphs}, whereas high-resource pairs train normally. This is interesting given that low-resource scores improved compared to bilingual baselines. Probably, the hierarchical model allows languages to learn from each other, but this is hindered by overfitting, and if we fix it, scores should grow even higher.

At the same time, this problem never arises in full sharing models (even when low-resource pairs are oversampled). There might be several reasons why this is happening. In hierarchical models, different translation paths have the same number of parameters regardless of the corpus size. This means that low-resource pairs may have too many parameters for the amount of data they have, hence, overfitting. And high-resource pairs may need more parameters than low-resource pairs. In this sense, full sharing models may be more effective, since the parameter space is distributed automatically. Another possible explanation is that in full sharing models high-resource data acts as an implicit regularizer, preventing model parameters from being overspecialized in a low-resource pair. This is not exactly so in hierarchical models---there are individual layers, and there can be layers shared by several close low-resource languages.

Yet another concern is oversampling low-resource data when training hierarchical models, which we applied and which itself can be the reason for overfitting. Indeed, we did not observe overfitting in our initial experiments without oversampling; However, in this case, low-resource scores are considerably lower than the ones we reported. Hence, we decided to fight overfitting using a different approach rather than removing oversampling. 

Based on this reasoning we suggest two ways to solve the problem of overfitting:

1) Downweight low-resource samples, i.e. decrease their weight in the loss function proportionally to data disbalance. Disbalance here is taken to be the ratio between the low-resource corpus size and the closest high-resource corpus size. For example, if the ratio is 1:5, then low-resource samples have five times less weight in the loss function. The idea is that this will act as a regularization for low-resource directions and at the same time high-resource samples will have more influence in shared parts which will allow improving high-resource scores too.

2) Regularize low-resource paths with high-resource data. This idea is based on the above hypothesis about an implicit regularization happening in full sharing models. Here we explicitly apply this regularization by feeding the closest high-resource data instead of the low-resource data intended for the path. I.e., during the epoch the low-resource path is fed all available low-resource samples (once, we do not oversample here) plus related high-resource samples. We downweight high-resource samples used for regularization proportionally to data disbalance to limit the possible negative effect.

We applied both these ideas. First of all, both of them solved the problem of overfitting. Now let us refer again to Figure~\ref{chart}. As demonstrated there, downweighting low-resource samples turned out to bring a lot of improvement. First, the model now performs on average better than both previous models (+0.42). Second, the issue with high-resource pairs is greatly alleviated, from -1.88 to -0.92. And third, low-resource scores have also noticeably increased, from +1.31 to +1.76. As for the second approach, it also improved high-resource scores (-1.09) but severely hit low-resource pairs. Probably, the reason is that the regularization was too strong, which prevented possible learning.

So, to summarize, both approaches solved the problem of overfitting, and while regularization with high-resource data turned out to be too strict, downweighting low-resource samples was able to greatly improve the scores for both high-resource and low-resource directions.

Now, taking the improved version of the hierarchical model as a primary one, we will try to answer the questions we put at the beginning of section~\ref{sec:experiments}. We were interested to know whether there is a difference when the hierarchy is on the encoder or decoder side. From our data we can say that there is no substantial difference: the hierarchical model performs at the same level regardless of where the hierarchy is. What is interesting though, simple hierarchical models perform considerably better than bigger ones: high-resource directions degrade less (-0.52 on average versus -1.33), and low-resource improve more (+2.02 versus +1.49). This may happen because adding more languages to a multilingual model inevitably leads to more distant pairs being mixed together, and they may interfere with each other instead of helping.  

Comparing hierarchical models to bilingual ones confirmed our supposition that hierarchical architecture allows languages to learn from each other, however, this is true mostly for low-resource pairs. If we look at the exact scores, high-resource directions do not always degrade: in 2 out of 8 cases they improve, although, not by a large margin. Low-resource directions improve in all 8 cases.

As for the comparison of hierarchical models with full sharing ones, the improved hierarchical model surpasses the full sharing model in both high-resource (+0.34) and low-resource (+0.48) pairs. At the same time, the full sharing model is beneficial in the sense that it comes with an intrinsic regularization mechanism, and perhaps this aspect could be further improved in hierarchical models.   

 \section{Conclusion}
\label{sec:conclusion}

This work tested a new hierarchical approach in multilingual MT. We implemented the hierarchical model based on the Transformer architecture and contrasted it to bilingual and full parameter sharing baselines. The straightforward approach to training turned out to be hampering the hierarchical model's performance. We found out that there is a problem of overfitting which is specific to training hierarchical models. Regularizing low-resource directions solved the problem, substantially improving the model's performance. 

We showed that the hierarchical model greatly improves low-resource pairs' scores, however, at the expense of high-resource pairs. The comparison with the full sharing model provided positive evidence supporting the assumption about the usefulness of explicitly defining parameter sharing strategy in a multilingual model. 


Overall, the hierarchical approach looks promising and hence should be further explored and developed. In the future, we would like to analyze how exactly learning happens in hierarchical models and whether hierarchical architecture is indeed capable of capturing different language aspects (morphological, syntactic, semantic) on different levels in a hierarchical manner.

\bibliography{eacl2021}
\bibliographystyle{acl_natbib}

\end{document}